\documentclass[letterpaper]{article} 
\usepackage{aaai2026}  
\usepackage{times}  
\usepackage{helvet}  
\usepackage{courier}  
\usepackage[hyphens]{url}  
\usepackage{graphicx} 
\usepackage{natbib}  
\usepackage{caption} 
\usepackage{algorithm}
\usepackage{algorithmic}
\usepackage{xspace}
\usepackage{booktabs}
\usepackage{latexsym}
\usepackage{inputenc}
\usepackage{microtype}
\usepackage{inconsolata}
\usepackage{tcolorbox}
\usepackage{xcolor}
\usepackage{amsmath}
\usepackage{amssymb}
\usepackage{multirow}
\usepackage{longtable}
\usepackage{array}
\usepackage{spverbatim}
\usepackage[table]{xcolor}

\usepackage{newfloat}
\usepackage{listings}
\DeclareCaptionStyle{ruled}{labelfont=normalfont,labelsep=colon,strut=off} 
\floatstyle{ruled}
\newfloat{listing}{tb}{lst}{}
\floatname{listing}{Listing}
\newcommand{\methodname}{HIPPO\xspace}
\newcommand{\contamination}{Pre-RL data overlap\xspace}
\newcommand{\Contamination}{Pre-RL Data Overlap\xspace}

\title{To Reason or to Fabricate: Reasoning Without Shortcuts \\via Hint-Anchored Pairwise Aggregation}
\author{
    Jiuheng Lin, 
    Chen Zhang and 
    Yansong Feng\thanks{Corresponding author.}
}
\affiliations{
    Wangxuan Institute of Computer Technology, Peking University\\
    linjiuheng@stu.pku.edu.cn \quad
    \{zhangch, fengyansong\}@pku.edu.cn
}

\usepackage{bibentry}

\begin{document}

\maketitle

\begin{abstract}

While reinforcement learning (RL) significantly enhances LLM reasoning, its efficacy is severely undermined by \textit{\contamination}, 
where RL datasets overlap with pretraining or SFT corpora, causing models to exploit shortcuts by memorizing correct answers and fabricating post-hoc reasoning. 
To address this, we introduce \methodname, a novel RL framework that integrates hint-injected aggregation with a tailored pairwise reward model. 
By utilizing hint injection to deliberately trigger overlap-induced behaviors, the resulting traces naturally serve as explicit anchors for pairwise comparison. This provides highly discriminable preference signals, enabling a lightweight judge model to reliably distinguish genuine reasoning deduction from shortcut-driven rationalization, while the pairwise formulation ensures stable and robust optimization compared to standard PRMs. 
Extensive experiments demonstrate that \methodname yields substantial improvements over standard baselines and generalizes effectively to out-of-distribution general tasks, 
showing it extracts authentic, transferable reasoning skills rather than superficial shortcut patterns.\footnote{Code is open sourced at: \url{https://github.com/Infinite-set/HIPPO} }

\end{abstract}


\section{Introduction}

\begin{figure}[t]
\centering
\includegraphics[scale=0.8]{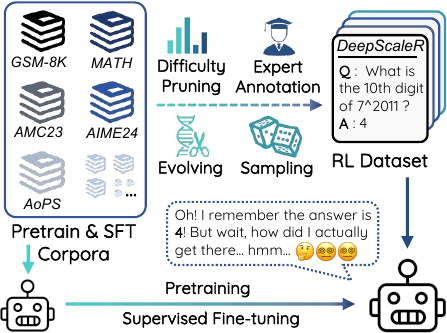}
\caption{An illustration of \contamination, using DeepScaleR~\cite{deepscaler2025} as an example. 
See Appendix~\ref{app:data_overlap} for detailed analysis across more RL datasets.}
\label{fig:introduction}
\end{figure}

Large Language Models (LLMs) have achieved remarkable success, with their reasoning capabilities recently experiencing massive leaps driven by Reinforcement Learning with Verifiable Rewards (RLVR)~\cite{grpo,guo2025deepseek,yu2025dapoopensourcellmreinforcement}. 
However, many commonly used challenging RL datasets inevitably overlap with the models' pretraining or supervised fine-tuning (SFT) corpora~\cite{wu2025reasoningmemorizationunreliableresults,balunović2026matharenaevaluatingllmsuncontaminated} due to reuse of scarce high-quality data pools, as illustrated in Figure~\ref{fig:introduction}. 
We formally define this phenomenon as \textit{\Contamination}.

\contamination severely undermines the efficacy of reinforcement learning. 
Under \contamination, the model retains partial memory of the solution without genuinely mastering the complete deductive process. 
While standard RLVR drives models to tackle complex tasks and autonomously discover reasoning patterns through final-answer correctness, 
exploiting superficial memory for answer retrieval also yields maximal rewards, 
acting as a trivial shortcut much easier to learn than executing rigorous logical deduction. 
Consequently, the training process degenerates into enhancing a retrieval shortcut—retrieving the memorized answer, then generating a post-hoc spurious rationale~\cite{lin2025clarityreasoningconsistencyteach,wu2025reasoningmemorizationunreliableresults}. Consequently, \contamination suppresses exploratory depth, degrades reasoning reliability and causing an illusory inflation of capabilities~\cite{yan2026spuriousrewardsparadoxmechanistically}.

Existing studies focus exclusively on detecting data overlap between training corpora and evaluation benchmarks~\cite{tao2026detectingdatacontaminationreinforcement,wang2026fragilitybenchmarkcontaminationdetection}, 
overlooking the negative impact of data overlap across training phases. 
Moreover, driven by the persistent demand to inject high-quality reasoning data into base models to establish strong reasoning capability boundaries and maximize their RL potential~\cite{yue2025doesreinforcementlearningreally,zhao2025echochamberrlposttraining,dong2026rlpluscounteringcapabilityboundary}, 
enforcing strict Pre-RL data isolation is both practically unfeasible and detrimental, as it inevitably discards valuable data and hinders model performance.
Therefore, we shift our focus from detection to mitigation and ask: \textit{How can we extract genuine reasoning gains from limited, partially contaminated datasets?}

To answer this, we propose \textbf{H}int-\textbf{I}njected \textbf{P}airwise \textbf{P}olicy \textbf{O}ptimization (\textsc{\methodname}), 
a novel RL framework designed to mitigate \contamination. 
We begin by formally modeling the active training policy and a hypothetical policy that has learned to exploit overlap-induced shortcuts.
The natural approach to penalize shortcut behaviors is to maximize the distributional KL divergence between them, 
but directly optimizing this divergence is mathematically intractable and unbounded. 
To overcome this, we theoretically derive a bounded surrogate objective by leveraging Pinsker's inequality and the dual form of TV distance~\cite{nguyen2010estimating}, 
finally reduce the intractable divergence into the reward expectation of generations produced by both policies. 

Next, we instantiate this hypothetical policy directly on-policy via answer-level hint injection. Since directly intervening in a model's internal parametric memory during Pre-RL overlap is difficult, this explicit hint serves as an accessible behavioral anchor. While operating on the input space, it effectively simulates the information-asymmetry of \contamination—knowing the answer before reasoning~\cite{chua2025deepseekr1reasoningmodels,chen2025reasoningmodelsdontsay}. By deliberately recreating this shortcut-reliant behavior~\cite{marioriyad2025unspokenhintsaccuracyacknowledgement,young2026liemefaithfulchainofthought}, we can algorithmically penalize it without requiring access to latent parametric states. Finally, we implement the reward judge as a pairwise Bradley--Terry comparator, maximizing the expected win rate of standard rollouts against hint-injected ones, thereby sidestepping the intransitivity and miscalibration that plague standard reward models~\cite{swamy2024minimaximalistapproachreinforcementlearning,xu2025investigatingnontransitivityllmasajudge,sutawika2026gainedtranslationprivilegedpairwise}.

Our experiments validate the effectiveness of \methodname across mathematical and medical reasoning. 
Compared to standard RL, \methodname yields consistent performance gains in both in-domain and out-of-distribution (OOD) scenarios, 
demonstrating capabilities to extract genuine reasoning from heavily overlapping datasets while maintaining strong generalizability. 
Furthermore, we show our hint-injected comparison provides a much clearer and more stable preference signal than direct pairwise comparison,
enabling cost-efficient small models to serve as reliable judges, 
and provide evidences for the mechanism by which \contamination degrades RL training. 

Our main contributions are summarized as follows:
(1) We identify the \contamination problem, theoretically reduce it to a bounded surrogate, and propose a mechanistic explanation regarding its impact on reasoning.
(2) We propose \methodname, a novel RL framework employing a hint-injected pairwise reward aggregation to suppress overlap-induced behaviors.
(3) Extensive experiments show that \methodname enhances reasoning and generalizability, while delivering superior optimization stability and clearer preference signals.

\section{Methodology}

\begin{figure*}[t]
\centering
\includegraphics[scale=0.75]{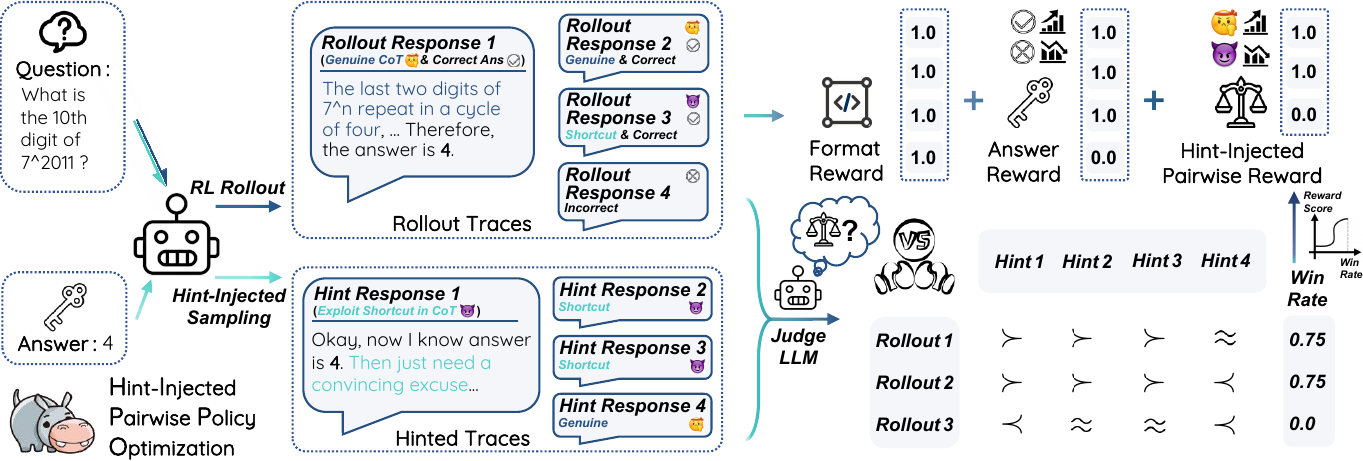}
\caption{The overall framework of \methodname. Given a task, \methodname performs standard RL rollouts paired with hint-injected sampling, followed by a pairwise comparison between the standard and the hint-injected traces.}
\label{fig:main_methodology}
\end{figure*}

In this section, we introduce \methodname, 
with its overall framework illustrated in Figure~\ref{fig:main_methodology}.
We begin by formally defining the \contamination mitigation task. 
Then we theoretically reduce the formulation into a bounded and empirically optimizable surrogate to overcome the intractable nature of directly optimizing divergence. 
Finally, we detail the practical implementation of our framework.

\subsection{Preliminary}

\paragraph{Reinforcement learning for LLMs.}
Let $(x, y) \sim \mathcal{D}$, where $x$ denotes a prompt and $y$ its reference answer.
A language model $\pi_\theta$ generates a reasoning trajectory $z \in \mathcal{Z}$ before emitting a final answer, with $\mathcal{Z}$ the space of token sequences.
Sampling proceeds autoregressively: 
$z \sim \pi_\theta(\cdot\mid x) = \prod_{t=1}^{|z|}\pi_\theta(z_t\mid x, z_{<t})$.
Under group-based RL training (e.g., GRPO and DAPO; ~\citealp{grpo,yu2025dapoopensourcellmreinforcement}), 
for each prompt $x$ we draw a group of $G$ rollouts $\{z^{(g)}\}_{g=1}^{G}\sim\pi_\theta(\cdot\mid x)$, score them with a scalar reward $r(\cdot)$, then form the standardized advantage and compute the PPO clipped surrogate~\citep{ppo}.

\paragraph{\contamination.}
\contamination---where the training data used for RL has already been exposed to the model during pre- or post-training---is prevalent and difficult to avoid in practice. 
To model and mitigate its effect, we formalize the trajectory distribution of the aforementioned contaminated shortcut-reliant hypothetical policy as $\pi_C(z \mid x)$.
This distribution characterizes a generation process where the model relies on memorized clues to bypass genuine deduction, thereby executing a reasoning shortcut directly leading to critical intermediate steps or ground-truth answer.


\subsection{Direct KL Optimization and Its Limitations}

The core training objective is to ensure that the current policy $\pi_\theta(z \mid x)$ remains distinguishable from $\pi_C(z \mid x)$ when trained on already contaminated datasets---meaning the policy actively avoids reasoning shortcuts even when it has implicitly memorized the answers due to \contamination.
A natural formalization is to maximize the Kullback--Leibler (KL) divergence between the two policies: 
\begin{equation}
D_{\mathrm{KL}}\!\bigl(\pi_\theta \big\| \pi_C\bigr) = \mathbb{E}_{z\sim\pi_\theta}\!\left[\log\frac{\pi_\theta(z\mid x)}{\pi_C(z\mid x)}\right]. \label{eq:kl_objective}
\end{equation}

Adopting Eq.~\eqref{eq:kl_objective} as a direct RL objective, however, is problematic in two ways. 
First, the per-sample divergence $\log[\pi_\theta/\pi_C]$ is a sum of token-level log-probability differences whose variance scales with trajectory length, 
overwhelming the policy gradient signal in long reasoning chains~\cite{yu2025dapoopensourcellmreinforcement,liu2025understandingr1zeroliketrainingcritical}. 
Second, the KL divergence is unbounded in $\theta$: $\pi_\theta$ can concentrate probability mass on regions where $\pi_c$ has near-zero density to make the divergence go to infinity, leading to reward hacking rather than genuine reasoning~\citep{gao2023scaling}.
These issues motivate replacing Eq.~\eqref{eq:kl_objective} with a stable and bounded surrogate.

\subsection{From KL Divergence to Pairwise Comparisons}
Directly optimizing the KL divergence is mathematically intractable. 
To overcome this, we introduce a reward-form surrogate estimated via pairwise comparisons between generations from active policy and contaminated policy. 
This mechanism provides a highly discriminable preference signal with superior robustness, tractability, and empirical estimability. 
We now formally derive how this pairwise approach serves as a provable lower bound surrogate for Eq.~\eqref{eq:kl_objective}.

\paragraph{Lemma 2.1 (Pinsker's inequality;~\citealp{tsybakov2009introduction,nguyen2010estimating}).}
\label{lem:pinsker}
For any probability distributions $p, q$ on $\mathcal{Z}$,
\begin{equation}
D_{\mathrm{KL}}(p\,\|\,q) \;\ge\; 2\,D_{\mathrm{TV}}^2(p, q),
\label{eq:pinsker}
\end{equation}
where $D_{\mathrm{TV}}(p, q) := \sup_{A\subseteq\mathcal{Z}}\bigl|p(A) - q(A)\bigr|. $ 

Pinsker's inequality furnishes a bounded surrogate lower bound for the unbounded KL objective, offering a safe guarantee for distancing the active policy from the shortcut-reliant behaviors.

\paragraph{Lemma 2.2 (Variational representation of TV).}
\label{lem:tv}
For any probability distributions $p, q$ on $\mathcal{Z}$ and any measurable function $T:\mathcal{Z}\to[0,1]$, the TV distance admits the equivalent form
\begin{equation}
D_{\mathrm{TV}}(p, q) \;=\; \sup_{T}\Bigl(\mathbb{E}_{z\sim p}[T(z)] - \mathbb{E}_{z\sim q}[T(z)]\Bigr),
\label{eq:tv-variational}
\end{equation}
with the supremum attained at $T^\star = \mathbf{1}_{A^\star}$, $A^\star := \{z\in\mathcal{Z}: p(z) \ge q(z)\}$.

This variational form converts a distance between intractable distributions into the expected score gap of a $[0,1]$-valued discriminator $T$, which is density-free and directly amenable to Monte Carlo estimation. Moreover, the supremum is attained precisely when $T$ correctly separates the two distributions, implying that the closer $T$ is to the optimal discriminator, the tighter the lower bound.

Concretely, in our setting, consider $p = \pi_\theta(\cdot\mid x)$ and $q = \pi_C(\cdot\mid x)$, for any capable reward model $T:\mathcal{Z}\to[0,1]$, 
\begin{equation}
\bigl(\mathbb{E}_{z\sim\pi_\theta}[T(z)] - \mathbb{E}_{z\sim\pi_C}[T(z)]\bigr)^2 \;\approx\; D_{\mathrm{TV}}^2(\pi_\theta, \pi_H).
\label{eq:lb}
\end{equation}
The leftmost term is a $[0,1]$-bounded expectation difference---density-free and estimable from rollouts---that provably lower-bounds the original KL objective. 
The next step is specifying $T$.

\paragraph{Instantiating $T$ via Pairwise Comparisons.}

The bound in Eq.~\eqref{eq:lb} is tight only when $T$ closely approximates the optimal discriminator $T^\star = \mathbf{1}_{A^\star}$, i.e., when $T$ accurately detects whether a trace is influenced by contamination. 
While tractable, evaluating a single trajectory via a pointwise reward model $T(z)$ can be problematic: 
previous works indicate that pointwise RM can only express transitive, Bradley--Terry-style preference structures and are notoriously poorly calibrated across distributions~\cite{xu2025investigatingnontransitivityllmasajudge,wang2026prefgrpopairwisepreferencerewardbased}. 

Pairwise comparators, by contrast, provably reduce variance and robustly handle non-Markovian or intransitive preference signals while being resilient to the compounding errors~\citep{swamy2024minimaximalistapproachreinforcementlearning,zhang2026arenarlscalingrlopenended}. 
We therefore instantiate $T$ via a pairwise judge $J: \mathcal{Z}\times\mathcal{Z}\to[0,1]$, where $J(z_a, z_b)$ denotes the judgment that $z_a$ is preferred to $z_b$. Concretely, we define
\begin{equation}
T_J(z) \;:=\; \mathbb{E}_{z'\sim\pi_C(\cdot\mid x)}\bigl[J(z, z')\bigr] \;\in\; [0, 1],
\label{eq:T_J}
\end{equation}
which automatically lies in $[0,1]$, satisfying the requirement of Lemma~\ref{lem:tv} and is agnostic to the implementation of $J$.

\paragraph{Lemma 2.3.}
\label{lem:sym}

Let $J$ be a pairwise comparator satisfying strict symmetry\footnote{In practice, we eliminiate the LLM judge's position bias by averaging both orderings; see Appendix~\ref{app:lemmas} for details.}, i.e., $J(z, z') + J(z', z) = 1$ for all $(z, z')$. Then for any distribution $q$, the marginal expectation over independent samples is constant:
\begin{equation}
\mathbb{E}_{z, z'\sim q}[J(z, z')] \;=\; \frac{1}{2}. \label{eq:symmetry}
\end{equation}

Substituting Eq.~\eqref{eq:T_J} into Eq.~\eqref{eq:lb} and applying Lemma~\ref{lem:sym} with $q = \pi_C$, the subtrahend collapses to a constant: $\mathbb{E}_{z\sim\pi_C}[T_J(z)] = \mathbb{E}_{z, z'\sim\pi_C}[J(z, z')] = \tfrac{1}{2}$.

\paragraph{Final Surrogate Objective.}

Combining Lemmas~\ref{lem:pinsker}--\ref{lem:sym}, our RL objective maximizes the following lower bound on the KL divergence:
\begin{equation}
\left|\mathbb{E}_{z\sim\pi_\theta,\,z'\sim\pi_C}\bigl[J(z, z')\bigr] - \frac{1}{2}\right|.
\label{eq:main_bound}
\end{equation}

Empirically, we observe that traces from the active policy $\pi_\theta$ typically exhibit higher quality than the contaminated baseline (i.e., $\mathbb{E}_{z\sim\pi_\theta,\,z'\sim\pi_C}[J(z, z')] \ge 1/2$; as detailed in Section~\ref{subsec:analysis_2}).
Therefore, maximizing the absolute value in Eq.~\eqref{eq:main_bound} is equivalent to maximizing the expectation alone, and we estimate it via an unbiased Monte Carlo reward:
\begin{equation}
r_{\mathrm{\methodname}}(z; x) := \frac{1}{K}\sum_{k=1}^{K} J\bigl(z, z'_{(k)}\bigr) \xrightarrow[]{K\to\infty} T_J(z),
\label{eq:r_hint}
\end{equation}
where $z'_{(k)} \sim \pi_C(\cdot\mid x)$ are $K$ independent trajectories sampled from the contaminated policy. 
This formulation completely reduces the pathological KL target into a bounded, variance-reduced stable pairwise signal. 
See more lemma prove and learning objective analysis in Appendix~\ref{app:lemmas}.

\subsection{Practical Instantiation of \methodname}

We detail the practical instantiation of the theoretical learning objective during RL training.

We first instantiate the contaminated shortcut-reliant policy $\pi_C$. 
Inspired by observations that prompt injections can reliably induce specific model behaviors (e.g., unfaithful reasoning or stylistic biases)~\cite{marioriyad2025unspokenhintsaccuracyacknowledgement,young2026liemefaithfulchainofthought}, we model the shortcut-reliant policy via an answer-level hint injection, denoted as $\pi_H(\cdot\mid x, h)$. 
Specifically, we inject the ground-truth answer $h$ into the prompt prior to the reasoning generation. 
This intervention has been empirically shown to effectively elicit the shortcut-reliant behaviors~\cite{marioriyad2025unspokenhintsaccuracyacknowledgement,young2026liemefaithfulchainofthought}.

Next, we instantiate the pairwise judge $J(z_a, z_b)$ by comparing standard rollouts against these hint-injected samples. 
As illustrated in Figure~\ref{fig:main_methodology}, for each rollout $z\sim\pi_\theta(\cdot\mid x)$ that yields a correct final answer, we sample $K$ hinted responses $z_h^{(k)}\sim\pi_H(\cdot\mid x, h)$ under the same model and conduct pairwise comparisons. 
Recalling that the supremum in Lemma~\ref{lem:tv} is attained when the reward function evaluates to a $\{0, 1\}$ binary score, we strictly binarize the continuous preference signal: win rate exceeding a predefined threshold ($0.5$) are mapped to $1$, while all others are strictly assigned $0$.

Finally, this pairwise reward is integrated with standard answer accuracy and format reward to form a comprehensive signal, 
penalizing flawed reasoning while suppressing shortcut-driven correct traces, 
ensuring that the policy is rewarded only for genuine, step-by-step deduction. Further details and discussions about reward configuration are provided in Appendix~\ref{app:impl} and ~\ref{app:more-exp}.

\section{Experimental Setup}
\label{sec:setup}

\paragraph{Implementation Details}

We evaluate \methodname on two representative domains: mathematical and medical reasoning. 

Training data is curated from DeepScaleR~\citep{deepscaler2025} for math and MedQA~\citep{jin2019pubmedqadatasetbiomedicalresearch} for medicine. 
We filter out trivial instances that the base model can solve in over 50\% of its trials, 
to ensure that the model has not yet mastered the genuine reasoning paths. 
This filtering yields about 5{,}000 instances for each domain. 
To deliberately simulate \contamination, 
ensuring the model acquires prior knowledge of the answer without the underlying deduction, 
we then conduct SFT using solely the final answers as targets, 
supplemented by a 20\% subset of data with rationales to prevent reasoning collapse. 
Finally we perform domain-specific RL training with the same data, to assess the performance under \contamination. 
We adopt \textbf{Qwen2.5-7B-Instruct} and \textbf{Qwen3-4B} as our backbone.


\paragraph{Evaluation Datasets}
We evaluate on both in-domain and out-of-domain (OOD) tasks to disentangle shortcut-driven gains from genuine reasoning improvements. 
For math, we evaluate on DeepScaleR~\citep{deepscaler2025}, MATH-500~\citep{lightman2023lets}, MMLU-Pro math subset~\citep{wang2024mmlu}, and CARP-EN~\citep{zhang2023evaluating}. 
For medicine, we utilize MedQA~\citep{jin2019pubmedqadatasetbiomedicalresearch}, MedMCQA~\citep{pmlr-v174-pal22a}, the MMLU-Pro Health and Biology subsets, and the GPQA Genetics and Molecular Biology subsets~\citep{rein2024gpqa}. 
OOD generalization is measured on TheoremQA~\citep{chen2023theoremqatheoremdrivenquestionanswering} together with the remaining MMLU-Pro subsets.
To mitigate randomness, we sample 4 responses per question and report the average accuracy.

\paragraph{Baselines}

To systematically disentangle the sources of performance gains and validate our theoretical reduction, 
we compare \methodname against two categories of baselines: 
(i) \textbf{Standard Training References} (\textit{SFT-only}, \textit{SFT with outcome-based RL}). These models bound the baseline contribution of the RL phase itself and quantify the gap between contamination-driven memorization and our shortcut-resistant training. 
(ii) \textbf{Reward-Shaping Alternatives}, including: \textit{RL with Pointwise RM} employs a standard absolute scalar reward model (standard PRM) to evaluate individual trajectory quality independently; 
\textit{Pref-GRPO}~\cite{wang2026prefgrpopairwisepreferencerewardbased} incorporates a pairwise preference framework, 
comparing and optimizing directly between all the rollout sequences; \textit{Privileged Pairwise Self-Play (SP3F)}~\citep{sutawika2026gainedtranslationprivilegedpairwise} further augments the pairwise judge model with privileged information (i.e., gold rationales). 
These baselines quantify the specific superiority of pairwise RM and our hint-injection mechanism.


\section{Results and Analysis}

\subsection{Main Results}
\label{subsec:main_exp}

We evaluate \methodname's effectiveness in Table~\ref{tab:math-result} and Table~\ref{tab:med-result} against all baselines across benchmarks. 

\noindent\textbf{Superior and Generalizable Performance.}
\methodname achieves consistent improvements over the base model and demonstrates superior performance against standard RL and all strong reward-shaping baselines. 
Moreover, the reasoning capabilities learned by \methodname are highly generalizable to broader, general-purpose tasks. 
Despite training on mathematical questions, our approach shows substantial gains on out-of-distribution general reasoning benchmarks. 

\noindent\textbf{Enhanced Reasoning Quality.}
Beyond accuracy, our approach elevates the response quality. As shown in Figure~\ref{fig:analysis:quality}, 
while standard RL degrades reasoning quality under \contamination, 
\methodname consistently produces more rigorous and coherent rationales across various backbone LLMs.

To fully unpack the mechanisms driving these gains, 
we structure our subsequent analyses by first validating our hint-injection paradigm as a reliable proxy for \contamination through fine-grained quality analysis, 
then demonstrating its substantial benefits on optimization stability and scaling, 
and finally mechanistically interpreting the reasoning degradation induced by \contamination.

\begin{table*}[t]
\centering
\small
\setlength\tabcolsep{6pt}
\begin{tabular}{ll|c|c|c|c|c|c}
\toprule
\multirow{2}{*}{\textbf{Method}} & & \multicolumn{3}{c|}{\textbf{Mathematical Reasoning}} & \multicolumn{2}{c|}{\textbf{General Reasoning}} & \multirow{2}{*}{\textbf{Avg.}} \\
\cmidrule{3-7}
& & \textbf{DeepScaleR} & \textbf{MATH-500} & \textbf{CARP-EN} & \textbf{TheoremQA} & \textbf{MMLU-Pro} &  \\
\midrule
\multirow{3}{*}{\textit{Standard Training}} & Qwen2.5$_\text{7B}$ & 10.1 & 12.1 & 41.9 & 28.3 & 33.0 & 25.1   \\
& \quad + SFT & 38.8 & 67.2 & 55.6 & 39.6 & 53.0 & 50.8 \\
& \quad \quad + RL & 45.2 & 68.2 & 58.7 & \underline{42.4} & 55.4 & \underline{54.0}     \\
 \midrule
\multirow{3}{*}{\textit{Reward-Shaping}} & PRM RL & 45.5 & 66.2 & 56.1 & 40.1 & 56.0 & 52.8   \\
& \textsc{Pref-GRPO} & 42.2 & \underline{68.3} & \underline{58.9} & 40.7 & \underline{56.8} & 53.4  \\
& SP3F & \underline{46.0} & 68.0 & 54.1 & 38.2 & 56.1 & 52.5 \\
\midrule
\rowcolor{gray!20}
\textit{Ours} & \textsc{\methodname} & \textbf{46.2} & \textbf{69.3} & \textbf{59.3} & \textbf{44.1} & \textbf{58.8} & \textbf{55.5} \\
\bottomrule
\end{tabular}
\caption{Performance of the model trained on DeepScaleR across mathematical tasks and out-of-domain general reasoning benchmarks. Best scores are \textbf{bold}, with the second \underline{underlined}. Results for Qwen3$_{\text{4B}}$ are detailed in Table~\ref{tab:app:qwen3-4b-math} in Appendix~\ref{app:more-exp}.}
\label{tab:math-result}
\end{table*}

\begin{table*}[t]
\centering
\small
\setlength\tabcolsep{8pt}
\begin{tabular}{l|c|c|cc|cc|c}
\toprule
\multirow{2}{*}{\textbf{Method}} & \multirow{2}{*}{\textbf{MedQA}} & \multirow{2}{*}{\textbf{MedMCQA}} & \multicolumn{2}{c|}{\textbf{MMLU-Pro}} & \multicolumn{2}{c|}{\textbf{GPQA}} & \multirow{2}{*}{\textbf{Avg.}} \\
 & & & Health & Biology & Genetics & MolBio &  \\
\midrule
Qwen2.5$_\text{7B}$ & 49.4 & 53.8 & 51.3 & 65.0 & 21.9 & 47.9 & 48.2  \\
\quad + SFT & 49.5 & 48.4 & 48.2 & 67.0 & 26.6 & 44.4 & 47.4 \\
\quad \quad + RL & \underline{61.2} & 55.1 & \textbf{57.5} & \underline{74.5} & 25.0 & 44.8 & \underline{53.0} \\
 \midrule
PRM RL & 57.7 & 52.0 & 41.1 & 65.7 & 21.8 & \underline{48.7} & 47.8  \\
\textsc{Pref-GRPO} & 56.6 & 51.8 & 52.9 & 71.7 & 25.0 & 41.1 & 49.9 \\
SP3F & 61.1 & \underline{56.0} & 55.4 & 68.0 & \underline{28.0} & 45.6 & 52.4  \\
\midrule
\rowcolor{gray!20}  
\textsc{\methodname} & \textbf{61.7} & \textbf{57.0} & \underline{56.8} & \textbf{75.0} & \textbf{32.8} & \textbf{48.8} & \textbf{55.4} \\
\bottomrule
\end{tabular}
\caption{Performance of the model trained on MedQA. Results for Qwen3$_{\text{4B}}$ are detailed in Table~\ref{tab:app:qwen3-4b-med} in Appendix~\ref{app:more-exp}.}
\label{tab:med-result}
\end{table*}

\subsection{Reasoning Degradation Analysis}
\label{subsec:analysis_1}

\begin{figure}[t]
\centering
\includegraphics[scale=0.35]{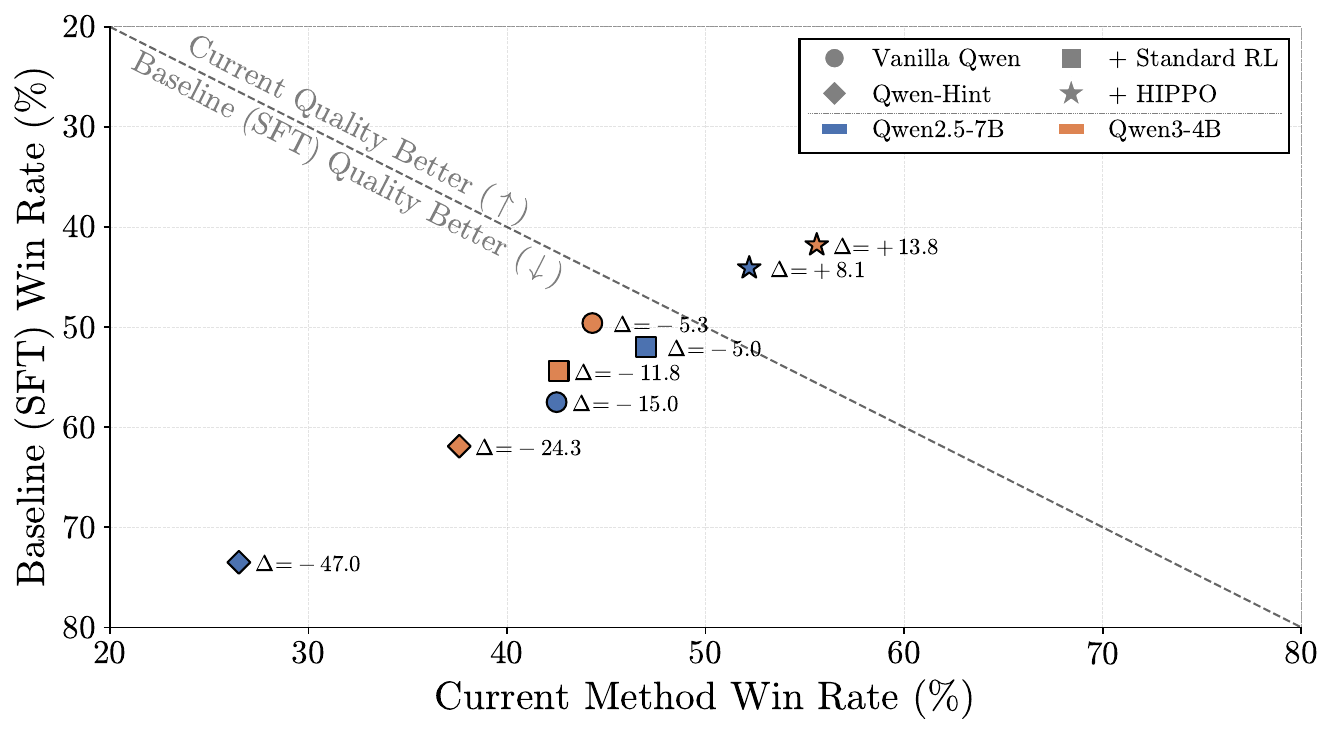}
\caption{Response quality of different methods on MedQA overlapped data, measured via pairwise comparisons against the SFT baseline using a strong judge (Gemini-3.1-Pro).}
\label{fig:analysis:quality}
\end{figure}

\paragraph{Under \contamination, standard RL degrades rationale quality.} We first investigate the direct impact of \contamination on reasoning capabilities. 
We sample correct responses using the MedQA \textit{training data} and conduct pairwise comparisons against responses from the SFT model to observe the performance shifts introduced by subsequent RL training. 
As illustrated in Figure~\ref{fig:analysis:quality}, applying standard RL notably yields lower rationale quality than the SFT baseline on these overlapping data. 
This degradation indicates that standard RL fails to enhance genuine reasoning (see more detailed mechanistic discussion on the RL-induced shortcut retrieval behavior in Section~\ref{sec:discussion}). 
In contrast, \methodname outperforms all baselines, demonstrating its capability to extract genuine reasoning rather than cultivating memorization.

\paragraph{The reliability of using hint-injection sampling as a behavior proxy for \contamination.} 
In Figure~\ref{fig:analysis:quality}, applying hint-injection substantially degrades the quality of generated responses, 
consistent with prior findings~\cite{chen2025reasoningmodelsdontsay,marioriyad2025unspokenhintsaccuracyacknowledgement,wang2026donttellanswertruly}. 
Beyond exhibiting a more severe drop in overall quality than that in standard RL under \contamination, 
we further delve into fine-grained quality characteristics to establish hint-injection as a more observable and comprehensive proxy.

As shown in Table~\ref{tab:analysis:quality}, using the correlation between reasoning error dimensions as an analytical lens, we find this correlation between most dimensions---measured by co-occurrence agreement---is significantly higher in our hint-injected paradigm than in standard RL failures. 
This indicates that \textit{errors are more coupled under hint-injection than natural \contamination}: rationales tend to be either faultless or simultaneously riddled with multiple flaws. 
Therefore, hint-injection amplifies the flaw density of natural \contamination and significantly improves observability—detecting solely a fraction of these errors suffices for the judge to identify poor rationales, 
thus lowering evaluation difficulty and enhancing discriminability, as empirically demonstrated next.

\begin{table}[t]
\centering
\small
\setlength\tabcolsep{4pt} 
\begin{tabular}{l|cccc}
\toprule
\textbf{Metric} & \textbf{Accuracy} & \textbf{Knowledge} & \textbf{Logic} & \textbf{Assertion} \\
\midrule
\textbf{Accuracy}    & - & 0.67 & 0.74 & 0.76 \\
\textbf{Knowledge}   & \textcolor{gray}{0.55} & - & 0.70 & 0.73 \\
\textbf{Logic}  & \textcolor{gray}{0.70} & \textcolor{gray}{0.66} & - & 0.84 \\
\textbf{Assertion} & \textcolor{gray}{0.75} & \textcolor{gray}{0.66} & \textcolor{gray}{0.84} & - \\
\bottomrule
\end{tabular}
\caption{Average quality scores across dimensions evaluated by Gemini-3.1-Pro. The upper-right and \textcolor{gray}{lower-left} triangles denote \textit{Hint-Injection} and \textcolor{gray}{\textit{Standard RL}}, respectively (detailed criteria in Appendix~\ref{app:impl}).}
\label{tab:analysis:quality}
\end{table}

\begin{table}[t]
\centering
\small
\setlength\tabcolsep{4pt}
\begin{tabular}{l|ccc}
\toprule
\textbf{Judge Model} & \textbf{Pointwise Acc} & \textbf{AUROC} & \textbf{Precision@1} \\
\midrule
Llama3.1$_\mathrm{8b}$ & 52.8 & 68.0 & 72.7 \\
Qwen2.5$_\mathrm{7b}$ & 53.9 & 73.2 & \textbf{86.9} \\
Qwen3$_\mathrm{4b}$   & \textbf{78.2} & \textbf{78.4} & 84.7 \\
\bottomrule
\end{tabular}
\caption{Performance comparison of different Judge models. Evaluation metrics include judge accuracy in pointwise assessment, alongside AUROC and Precision@1 in pairwise evaluation (we set $N=8$).}
\label{tab:judge-model-performance}
\end{table}

\subsection{Why Hint-Injected Pairwise Rewards Work}
\label{subsec:analysis_2}

\paragraph{Efficacy of Small Open-Source LLMs in Pairwise Evaluation.} 
To validate the reliability of small pairwise LLM judges, 
we sample correct reasoning traces together with their hinted counterparts on the validation set,  
and use a strong LLM (Gemini-3.1-pro) to establish ground-truth quality annotations. 
we then evaluate judge of varying size and architecture. 
As shown in Table~\ref{tab:judge-model-performance}, pointwise evaluation heavily relies on the capability of the judge model. Without sufficient capacity, small models degrade to near-random guessing, which explicitly explains the poor performance of our pointwise RM baseline. 
In contrast, pairwise judgment demonstrates superior stability across all evaluated models, consistently achieving high AUC-ROC and Precision@1 scores, 
which shows its effectiveness in both global preference ranking and precise top-quality trace identification. 
These properties are valuable for group-based RL optimization, as it helps provide a clear and stable gradient signal for policy alignment.

\paragraph{Hint-Injection Enhances Discriminability and Transitivity.}
By serving as a contaminated reference anchor, the hinted reference simplifies the evaluation landscape, yielding more stable and consistent optimization.

\begin{figure}[t]
\centering
\includegraphics[scale=0.283]{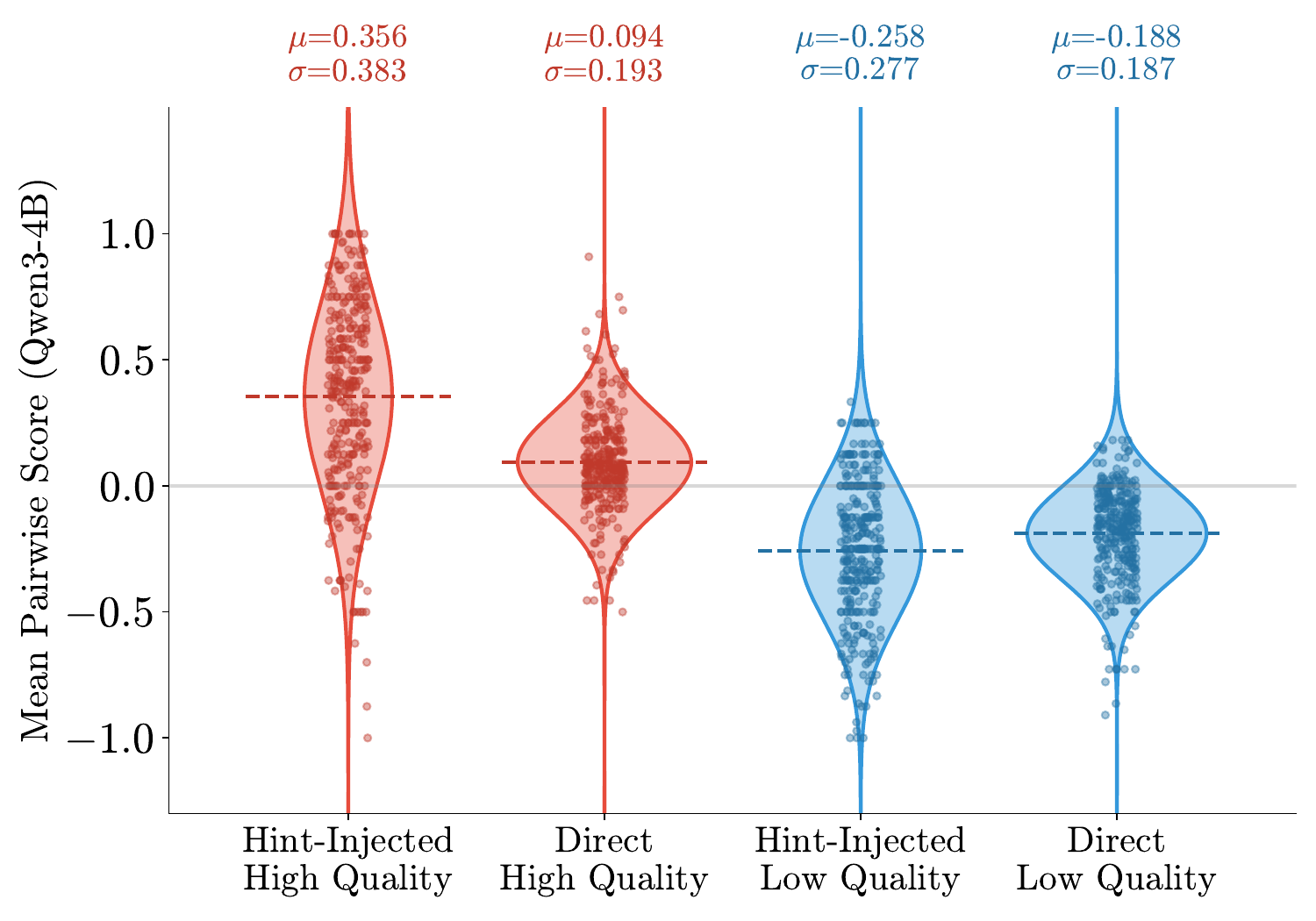}
\caption{Score comparison between the direct and hint-injected pairwise judges. Hint-injection effectively enlarges the decision boundary between high- and low-quality traces.}
\label{fig:analysis:contrast}
\end{figure}

First, it sharpens the contrast between high- and low-quality traces. 
We examine this by comparing the pairwise scores derived from our hint-injected approach against a standard pairwise baseline (i.e., comparing pairs of vanilla rollout generations directly). 
As shown in Figure~\ref{fig:analysis:contrast}, 
the average distance between ground-truth high- and low-quality traces increases substantially under hint-injection, 
yielding a clearer separable boundary and a more informative reward signal.

\begin{figure}[t]
\centering
    \includegraphics[scale=0.222]{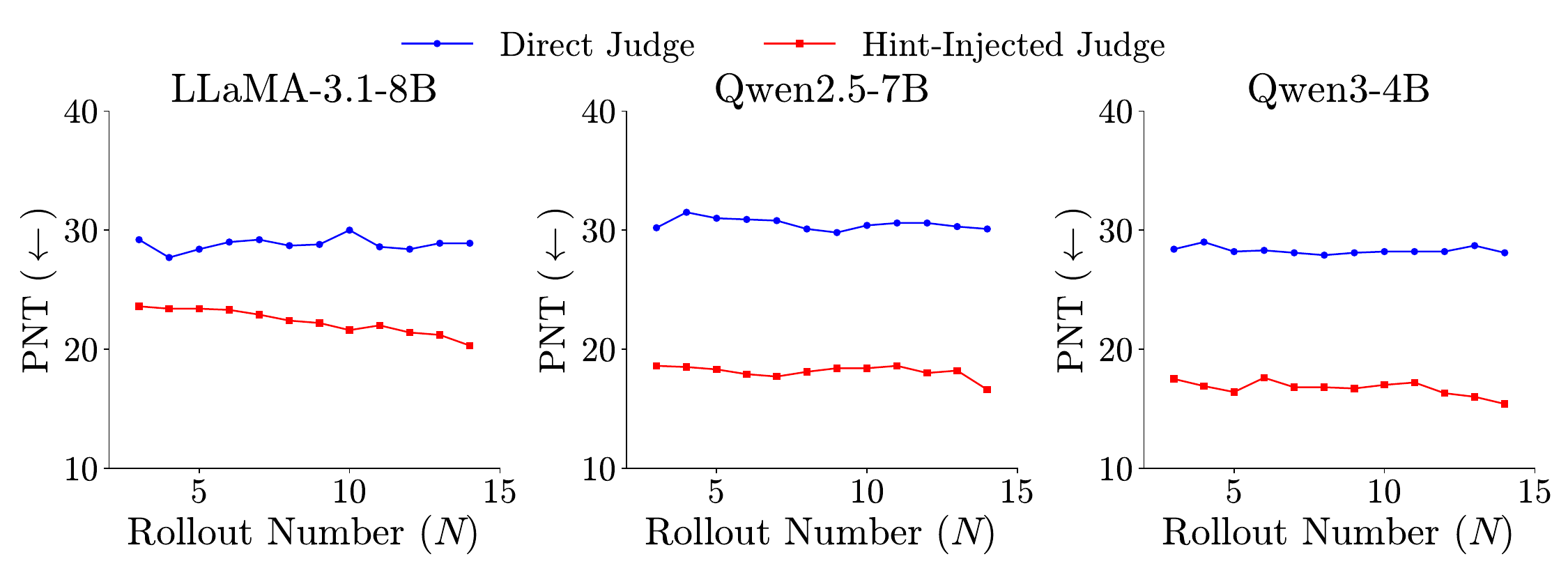}
\caption{Comparison of PNT statistics between the direct pairwise judge and our hint-injected pairwise judge. Our proposed method consistently exhibits better evaluation stability and overall performance across all tested models.}
\label{fig:analysis:PNT}
\end{figure}

Second, hint-injected pairwise judging significantly mitigates the issue of \textit{intransitive preferences}. 
As noted by previous works~\cite{swamy2024minimaximalistapproachreinforcementlearning,xu2025investigatingnontransitivityllmasajudge}, pointwise reward models are notoriously prone to it. 
While switching to pairwise comparison alleviates this issue~\cite{sutawika2026gainedtranslationprivilegedpairwise,wang2026prefgrpopairwisepreferencerewardbased}, our hint-injected approach further maximizes this mitigation. 
We quantify this issue via the Percentage of Non-Transitive  (PNT) metric~\citep{xu2025investigatingnontransitivityllmasajudge}---the fraction of trace triplets that violate transitivity. 
Figure~\ref{fig:analysis:PNT} demonstrates that our hint-injected approach consistently achieves lower PNT than direct pairwise comparisons, 
thus substantially simplifying the preference learning problem. 
Moreover, our method exhibits a distinctly positive scaling trend---as model capacity increases, our approach continuously drives the PNT lower without bottlenecking, demonstrating significantly superior effectiveness and scalability.
See Appendix~\ref{app:pnt} for calculation details.


\section{Exploring the Degradation Mechanism}
\label{sec:discussion}
While the preceding analyses establish that \contamination degrades reasoning reliability and generalizability, 
its driving mechanism remains underexplored. 
We thus propose a hypothesis: 
\textit{familiar questions trigger memorized answer retrieval rather than genuine deduction.}
Consequently, RL optimizes for memory exploitation over reasoning enhancement. 
In this section, we validate this hypothesis by demonstrating that \contamination indeed triggers premature answer retrieval, reducing the subsequent reasoning to mere post-hoc rationalizations.

\paragraph{Experimental Setup.} 
We design two pilot experiments. 
First, we examine the internal representations of the model under \contamination using the Logit Lens ~\cite{nostalgebraist2020logitlens,belrose2025elicitinglatentpredictionstransformers}. 
By directly projecting intermediate hidden states from all layers into the vocabulary space across the entire generation sequence, 
we verify whether the final answer token becomes identifiable at shallow layers or during the early beginning of token generation. 
Either scenario implies that the target answer is accessible prior to substantial reasoning, signaling a shortcut.

Second, we employ the Independent Causal Mechanisms (ICM) principle~\citep{10.1016/j.artint.2012.01.002,schoelkopf2012causalanticausallearning,jin-etal-2021-causal} to analyze the generative information flow under overlap. 
ICM postulates that the true causal direction $C \rightarrow E$ minimizes Kolmogorov complexity, satisfying $K(P_C) + K(P_{E|C}) \le K(P_E) + K(P_{C|E})$, which means the non-causal direction contains redundant information.
Since true Kolmogorov complexity is uncomputable, we follow previous works~\citep{burden2024conversationalcomplexityassessingrisk} and approximate $K(\cdot)$ using negative token log-likelihoods. 
We evaluate the complexity difference between the two generative directions---$[K(y) + K(z \mid y)] - [K(z) + K(y \mid z)]$---across the base, SFT, and SFT+RL models, 
to reveal the causal direction shifts the reasoning $z$ and the final answer $y$.
Detailed theoretical justifications and experimental settings are provided in Appendix~\ref{app:causal}.

\begin{figure}[t]
\centering
    \includegraphics[scale=0.24]{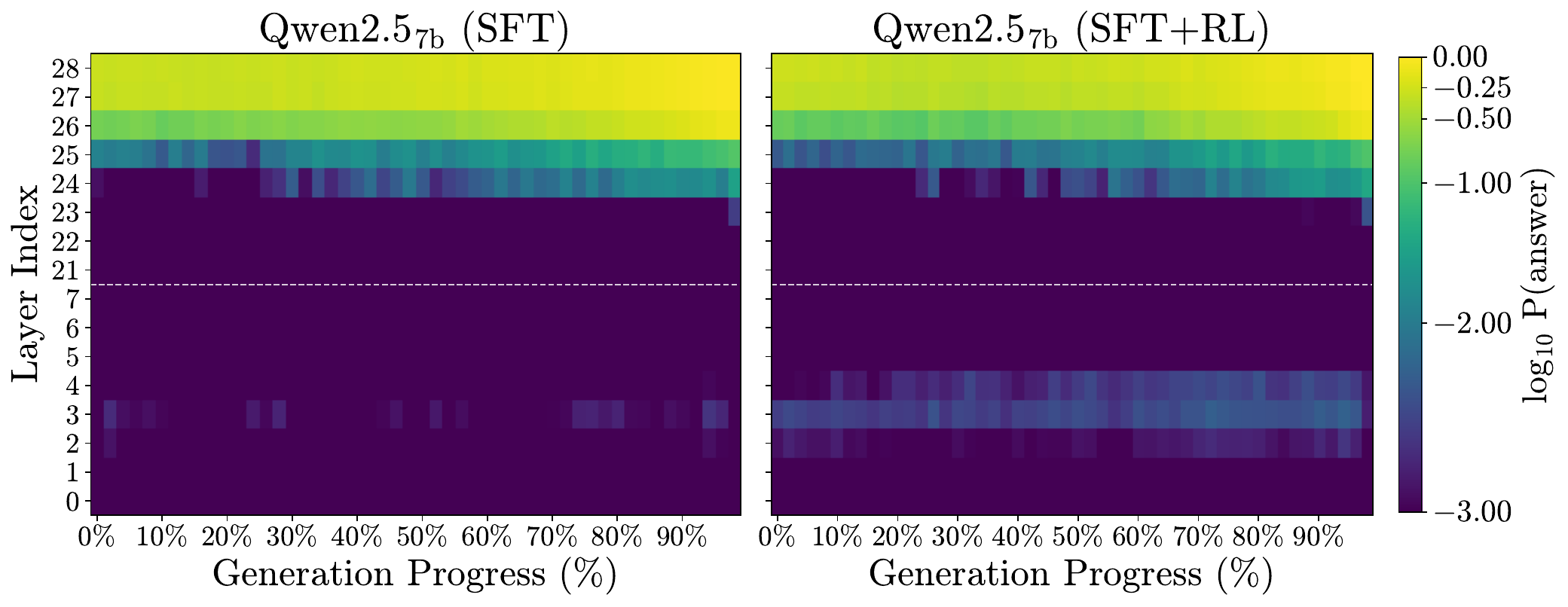}
\caption{Logit Lens visualizations of answer probability across hidden layers and generation steps.}
\label{fig:discussion:probing}
\end{figure}


\begin{table}[t]
\centering
\small
\setlength\tabcolsep{2.2pt}
\begin{tabular}{l|ccc|ccc}
\toprule
\multirow{2}{*}{\textbf{Model}} & \multicolumn{3}{c|}{\textit{MATH-500}} & \multicolumn{3}{c}{\textit{DeepScaleR}} \\
\cmidrule(lr){2-7}
& \textbf{A+R\textbar A} & \textbf{R+A\textbar R} & $\boldsymbol{\Delta}$ & \textbf{A+R\textbar A} & \textbf{R+A\textbar R} & $\boldsymbol{\Delta}$ \\
\midrule
Qwen2.5$_\mathrm{7B}$ & 63.22 & 49.50 & 13.72 & 71.42 & 56.58 & 14.84 \\
\quad + SFT           & 96.80 & 89.88 & 6.92  & 104.86 & 98.78 & 6.08 \\
\quad \quad + RL      & 90.45 & 86.17 & 4.28  & 90.85  & 89.92 & \textbf{0.93} \\
\bottomrule
\end{tabular}
\caption{Estimated complexity on overlapping data (DeepScaleR training set) compared to a uncontaminated dataset (MATH-500).}
\label{tab:contamination_casual_results}
\end{table}

\paragraph{Results}
The Logit Lens visualizations in Figure~\ref{fig:discussion:probing} reveal a behavioral divergence after the model undergoes RL. 
Specifically, we observe a dramatic increase in the target answer's probability within exceptionally \textit{shallow hidden layers}. 
This phenomenon indicates that the model forms an internal representation of the final answer well before executing any meaningful latent computation.
Moreover, this shallow-layer prominence persists throughout the entire generation process, 
providing clear evidence that the model anchors its output to a pre-retrieved answer rather than deriving it sequentially. 
This observation challenges the authenticity and functional necessity of the generated reasoning chain.

The result of causal information flow, shown in Table~\ref{tab:contamination_casual_results}, further confirms the model's already known the answer before reasoning.  
On the non-overlapping MATH-500 dataset, the causal direction remains robustly aligned with $z \rightarrow y$, as evidenced by the large positive $\Delta$ values across all training stages, indicating that reasoning naturally precedes the conclusion. 
However, on the overlapping DeepScaleR data, this complexity gap collapses precipitously after RL training, dropping to almost zero ($\mathbf{0.93}$). 
This structural collapse reflects a shift from a strictly forward reasoning process toward a mixed causal dynamic: 
while the model still genuinely resolves half of the instances, 
\textit{the other half now clearly exhibits a reversed causal flow}:
the model anchors on the answer-retrieval shortcut, relegating the subsequent chain-of-thought to a fabricated, post-hoc rationalization. 
Together, these empirical findings validate our hypothesis, demonstrating that \contamination inherently drives RL to replace sequential logical deduction with shortcut-driven memory retrieval.

\section{Related Works}

\paragraph{RL for LLM Reasoning}
While RL  has driven remarkable leaps in LLM reasoning, it carries inherent risks. 
Without explicit reasoning monitoring, RL can stagnate performance or even induce detrimental behaviors, 
such as logical inconsistency~\cite{lin2025clarityreasoningconsistencyteach}, unfaithfulness~\cite{chen2025reasoningmodelsdontsay}, or collapsing into mere memory retrieval due to reward breakdown~\cite{wu2025reasoningmemorizationunreliableresults}. 
To monitor the reasoning process, various Process Reward Models have been proposed. 
Although large-scale commercial models deliver state-of-the-art evaluation, they incur prohibitive costs~\cite{chen2024huatuogpto1medicalcomplexreasoning,zhang2026arenarlscalingrlopenended}. 
Alternatively, optimizing the evaluation paradigm can enable much smaller LLMs to serve as PRMs. 
Pairwise reward modeling, a prominent advancement in this vein, effectively addresses the instability, intransitivity, and miscalibration that plague standard PRMs~\cite{swamy2024minimaximalistapproachreinforcementlearning}. Furthermore, it better aligns with human annotation behavior by evaluating relative preferences between ambiguous trajectories~\cite{wang2026prefgrpopairwisepreferencerewardbased,zhang2026arenarlscalingrlopenended}. 

Building upon this pairwise framework, our method introduces hint-injected generation as an explicit anchor. This design further stabilizes the optimization process and yields clearer, more discriminable reward signals, particularly under the challenge of \contamination.

\paragraph{Data Overlap Across Training Stages}
Previous studies mostly adopt a \emph{training--evaluation} lens, focusing on whether training corpora leak into test benchmarks~\cite{cheng2025surveydatacontaminationlarge}. 
Within the context of pre-training and SFT, this perspective has driven the development of output-distribution detectors~\cite{dong-etal-2024-generalization}, paraphrase-evasive probes~\cite{lan2026illusionreasoningexposingevasive}, and competition-math audits~\cite{balunović2026matharenaevaluatingllmsuncontaminated}, alongside stress tests revealing that such detection mechanisms often falter on modern reasoning models~\cite{wang2026fragilitybenchmarkcontaminationdetection}. 
Other works extend this paradigm to RL training, investigating whether exposure to specific RL-stage data spuriously inflates benchmark scores~\cite{tao2026detectingdatacontaminationreinforcement,wu2026positionhiddencostsmeasurement,kocyigit2026impactposttrainingdatacontamination}. While existing analyses remain rigidly anchored to evaluation leakage, we identify a critical yet underexplored regime where pre-RL corpora act as the contamination source, with the RL stage itself serving as the vulnerable sink.

\section{Conclusion}
We identify the \contamination problem which severely undermines RL efficacy by driving models to exploit memorization shortcuts. 
We introduce \methodname, a novel framework that theoretically reformulates an intractable divergence objective into a stable, pairwise-rewarded surrogate. 
By utilizing hint-injected aggregation to deliberately trigger shortcut behaviors, 
the resulting traces naturally serve as explicit anchors for pairwise comparison. 
\methodname yields a highly discriminable, stable, and learnable preference signal, ultimately achieving superior reasoning capabilities and strong generalizability.

\bibliography{aaai2026}

\appendix

\clearpage

\section{Proofs}
\label{app:lemmas}

\subsection{Proof of Lemma~\ref{lem:pinsker} (Pinsker's inequality)}
\label{app:pinsker}

We prove $D_{\mathrm{KL}}(p\|q) \ge 2\,D_{\mathrm{TV}}^2(p,q)$ in two steps: (i) reduction to a Bernoulli case via the data-processing inequality, and (ii) direct verification for Bernoulli distributions.

\paragraph{Step (i): Reduction.}
Let $A^\star = \{z : p(z) \ge q(z)\}$ and define $f(z) := \mathbf{1}\{z \in A^\star\}$. The pushforward distributions $P := f_\# p$ and $Q := f_\# q$ are Bernoulli on $\{0,1\}$: $P = \mathrm{Ber}(\alpha)$, $Q = \mathrm{Ber}(\beta)$ with $\alpha := p(A^\star)$, $\beta := q(A^\star)$.

By the data-processing inequality for KL divergence, $D_{\mathrm{KL}}(p\|q) \ge D_{\mathrm{KL}}(P\|Q)$.
Meanwhile, $D_{\mathrm{TV}}(P, Q) = |\alpha - \beta| = p(A^\star) - q(A^\star) = D_{\mathrm{TV}}(p, q)$, where the last equality follows from Lemma~\ref{lem:tv} with the optimal witness $T^\star = \mathbf{1}_{A^\star}$.
Hence it suffices to prove $D_{\mathrm{KL}}(P\|Q) \ge 2\,D_{\mathrm{TV}}^2(P, Q)$ for the Bernoulli case.

\paragraph{Step (ii): Bernoulli case.}
We need to show that for all $\alpha, \beta \in (0,1)$,
\begin{equation}
\small
g(\alpha, \beta) \;:=\; \alpha\log\frac{\alpha}{\beta} + (1-\alpha)\log\frac{1-\alpha}{1-\beta} - 2(\alpha - \beta)^2 \;\ge\; 0.
\label{eq:pinsker-bernoulli}
\end{equation}
Fix $\beta$ and regard $g$ as a function of $\alpha$. Then:
\begin{itemize}
\item $g(\beta, \beta) = 0$.
\item $\dfrac{\partial g}{\partial \alpha}\bigg|_{\alpha=\beta} = \log\dfrac{\beta}{\beta} - \log\dfrac{1-\beta}{1-\beta} - 4(\beta - \beta) = 0$.
\item $\dfrac{\partial^2 g}{\partial \alpha^2} = \dfrac{1}{\alpha} + \dfrac{1}{1-\alpha} - 4 = \dfrac{1}{\alpha(1-\alpha)} - 4 \;\ge\; 4 - 4 = 0$, where the inequality uses $\alpha(1-\alpha) \le 1/4$.
\end{itemize}
Since $g$ is convex in $\alpha$ with $g(\beta,\beta) = 0$ and $\partial_\alpha g|_{\alpha=\beta} = 0$, we conclude $g(\alpha, \beta) \ge 0$ for all $\alpha \in (0,1)$. \hfill$\square$

\subsection{Proof of Lemma~\ref{lem:tv} (Variational representation of TV)}
\label{app:tv}

We show $D_{\mathrm{TV}}(p,q) = \sup_{T:\mathcal{Z}\to[0,1]}\bigl(\mathbb{E}_p[T] - \mathbb{E}_q[T]\bigr)$.

\paragraph{Upper bound.}
For any $T:\mathcal{Z}\to[0,1]$,
\begin{equation}
\mathbb{E}_p[T] - \mathbb{E}_q[T] = \sum_z T(z)\bigl(p(z) - q(z)\bigr).
\end{equation}
Partition the sum over $A^\star := \{z : p(z) \ge q(z)\}$ and $A^{\star c} := \{z : p(z) < q(z)\}$:
\begin{equation}
\begin{split}
\mathbb{E}_p[T] - \mathbb{E}_q[T] = \underbrace{\sum_{z\in A^\star} T(z)\bigl(p(z) - q(z)\bigr)}_{\ge\, 0} \\
+ \underbrace{\sum_{z\in A^{\star c}} T(z)\bigl(p(z) - q(z)\bigr)}_{<\, 0}.
\end{split}
\end{equation}
Applying $T(z) \le 1$ on $A^\star$ and $T(z) \ge 0$ on $A^{\star c}$:
\begin{equation}
\begin{split}
\mathbb{E}_p[T] - \mathbb{E}_q[T] \;\le\; \sum_{z\in A^\star}\bigl(p(z) - q(z)\bigr) \\
= p(A^\star) - q(A^\star) \le\; D_{\mathrm{TV}}(p, q),
\end{split}
\end{equation}
where the last inequality follows from the definition $D_{\mathrm{TV}}(p,q) = \sup_A |p(A) - q(A)|$.

\paragraph{Achievability.}
Setting $T = \mathbf{1}_{A^\star}$ yields $\mathbb{E}_p[\mathbf{1}_{A^\star}] - \mathbb{E}_q[\mathbf{1}_{A^\star}] = p(A^\star) - q(A^\star)$.
By the Hahn decomposition, $D_{\mathrm{TV}}(p,q) = p(A^\star) - q(A^\star)$, so the supremum is attained. \hfill$\square$

\subsection{Proof of Lemma~\ref{lem:sym} (Symmetry pins the baseline)}
\label{app:sym}

Under $q \otimes q$, the variables $z$ and $z'$ are exchangeable, so
\begin{equation}
\begin{split}
\mathbb{E}_{z,z'\sim q}[J(z, z')] = \mathbb{E}_{z,z'\sim q}[J(z', z)] \\
= \mathbb{E}_{z,z'\sim q}[1 - J(z, z')] \\
= 1 - \mathbb{E}_{z,z'\sim q}[J(z, z')],
\end{split}
\end{equation}
whence $\mathbb{E}_{z,z'\sim q}[J(z, z')] = 1/2$. \hfill$\square$

\paragraph{Practical realization.}
LLM-as-judge implementations of $J$ exhibit non-negligible position bias and therefore violate $J(z, z') + J(z', z) = 1$ in general. We restore symmetry at the implementation level by averaging the two orderings,
\begin{equation}
\tilde{J}(a, b) \;:=\; \tfrac{1}{2}\bigl(J(a, b) + 1 - J(b, a)\bigr),
\label{eq:J_sym}
\end{equation}
which satisfies $\tilde{J}(a, b) + \tilde{J}(b, a) = 1$ by construction. All downstream identities are stated for the symmetrized comparator $\tilde{J}$; for readability we write $J$ in place of $\tilde{J}$ hereafter.

\subsection{More Analysis on Learning Objectives}
\label{app:variance}


We compare the variance of the pairwise reward $r_{\mathrm{\methodname}}$ (Eq.~\eqref{eq:r_hint}) with that of the per-sample KL reward $r^*(z) = \log[\pi_\theta(z\mid x)/\pi_H(z\mid x, h)]$.

\paragraph{KL reward: variance grows with trajectory length.}
Under autoregressive decomposition, $r^*(z) = \sum_{t=1}^T \Delta_t$ where $\Delta_t = \log\frac{\pi_\theta(z_t\mid x, z_{<t})}{\pi_H(z_t\mid x, h, z_{<t})}$ and $T = |z|$. When the token-level log-ratios have a common positive variance lower bound $\sigma^2 > 0$ and non-negative correlations---a reasonable assumption when $\pi_\theta$ and $\pi_H$ diverge consistently across steps---we have $\mathrm{Var}[r^*] \ge T\sigma^2$, scaling at least linearly with trajectory length~\citep{schulman2018highdimensionalcontinuouscontrolusing}.

\paragraph{Pairwise reward: variance is uniformly bounded.}
Since $J \in [0,1]$, the reward $r_{\mathrm{\methodname}} \in [0,1]$ as well. By Popoviciu's inequality, any random variable supported on $[a,b]$ satisfies $\mathrm{Var}[X] \le (b-a)^2/4$, so
\begin{equation}
\mathrm{Var}[r_{\mathrm{\methodname}}] \;\le\; \frac{1}{4},
\end{equation}
independent of trajectory length $T$. More precisely, by the law of total variance,
\begin{equation}
\mathrm{Var}[r_{\mathrm{\methodname}}] = \underbrace{\mathbb{E}_{z}\!\bigl[\mathrm{Var}_{\{z\}}[r_{\mathrm{div}} \mid z]\bigr]}_{\le\; 1/(4K)} + \underbrace{\mathrm{Var}_{z}[T_J(z)]}_{\le\; 1/4}.
\end{equation}
The first term is the Monte Carlo estimation noise, which vanishes as $K \to \infty$; the second is the irreducible variance due to the variability of $z$ under $T_J$, bounded by $1/4$ since $T_J \in [0,1]$.

\section{Evaluation Details}
\label{app:eval}

\subsection{Percentage of Non-Transitive cases (PNT)}
\label{app:pnt}

We follow the same set of non-transitivity criteria as~\citet{xu2025investigatingnontransitivityllmasajudge}; the complete enumeration of violation patterns is given in their appendix and we do not reproduce it here.

For the direct (unhinted) baseline, we apply this protocol verbatim: every triple of rollouts drawn for the same prompt is compared by the judge model, and the fraction of triples that exhibit any of the violation patterns is reported as PNT.

In our hint-injected setting, the protocol is adapted on a single axis: the middle element of each triple is a \emph{hinted} rollout rather than an unhinted one. The two edges involving the hinted reference are still decided by the judge model, but the relative quality between the two outer (unhinted) rollouts is determined by their pairwise scores against the hint set rather than by a separate judge query. All other counting rules follow~\citet{xu2025investigatingnontransitivityllmasajudge} unchanged.

\subsection{Gold Quality Score}
\label{app:gold-score}
Throughout the analyses in Section~\ref{subsec:analysis_1}, Section~\ref{subsec:analysis_2} and Appendix~\ref{app:pnt}, the gold scalar score $r(z)$ is obtained by prompting a stronger judge model (Gemini-3.1-pro) to assign a $[0,1]$ score on each of a fixed set of quality dimensions. A trace is labeled \emph{good} only if it receives a passing score on every dimension; otherwise it is labeled \emph{bad}. The full prompt template is documented in our released code.

\subsection{Causal Analysis: Detailed Setup}
\label{app:causal}

This section provides the formal background and the experimental protocol omitted from Section~\ref{sec:discussion}.

\paragraph{Independent causal mechanisms.}
The diagnostic in Section~\ref{sec:discussion} rests on the principle of \emph{independent causal mechanisms} (ICM)~\citep{10.1016/j.artint.2012.01.002,schoelkopf2012causalanticausallearning,jin-etal-2021-causal}, which posits that the causal generative process of a system's variables decomposes into autonomous modules that carry no shared information. A direct corollary is that, for any pair of variables $(C, E)$ standing in a true causal relation $C \rightarrow E$, the Kolmogorov complexity of the joint distribution admits the asymmetric factorization
\begin{equation}
\small
K(P_{C,E}) \;=\; K(P_C) + K(P_{E \mid C}) \;\le\; K(P_E) + K(P_{C \mid E}),
\label{eq:icm}
\end{equation}
because the causal factorization is built from two independent modules ($P_C$ and the mechanism $P_{E \mid C}$), whereas the anti-causal factorization mixes them and typically carries redundant information. The sign of the gap $\Delta := \bigl[K(P_E) + K(P_{C \mid E})\bigr] - \bigl[K(P_C) + K(P_{E \mid C})\bigr]$ therefore identifies the causal direction.

\paragraph{Token-probability surrogate.}
Kolmogorov complexity is uncomputable. Following~\citet{burden2024conversationalcomplexityassessingrisk}, we approximate it by the negative log-likelihood under the policy:
\begin{equation}
K(t \mid \text{context}) \;\approx\; -\sum_{i=1}^{|t|} \log p_\theta\bigl(t_i \,\big|\, t_{<i}, \text{context}\bigr).
\label{eq:k-surrogate}
\end{equation}
Substituting Eq.~\eqref{eq:k-surrogate} into Eq.~\eqref{eq:icm} with $C$ instantiated by the reasoning trace $R$ and $E$ by the answer $A$, we obtain the empirical estimator
\begin{equation}
\small
\widehat{\Delta}(x) \;=\; \bigl[K(A \mid x) + K(R \mid A, x)\bigr] - \bigl[K(R \mid x) + K(A \mid R, x)\bigr],
\end{equation}
which is positive when the data are best explained by $R \rightarrow A$ and negative under $A \rightarrow R$.

\paragraph{Experimental protocol.}
For each policy under inspection (Vanilla, SFT-only, SFT-then-RL) and each evaluation set (contaminated, OOD), we sample $n$ rollouts per prompt, retain only those whose final answer is correct, and compute $\widehat{\Delta}(x)$ on each retained instance. The per-instance gaps are then averaged within the eligible subset, so that the reported $\Delta$ in Table~\ref{tab:contamination_casual_results} represents the mean causal-direction gap conditional on answer correctness. Restricting to correct samples removes the confound that incorrect traces may exhibit artificially low $K(A \mid R)$ for reasons unrelated to causal structure.

\begin{table}[t]
\centering
\small
\setlength\tabcolsep{6pt}
\begin{tabular}{lc}
\toprule
\textbf{Reward Variant} & \textbf{MedQA (val)} \\
\midrule
Soft reward (raw win-rate)        & 58.3 \\
Hard reward ($\alpha_{\mathrm{div}}=0.125$) & 62.0 \\
Hard reward ($\alpha_{\mathrm{div}}=0.25$) & 61.7 \\
\bottomrule
\end{tabular}
\caption{Hard binarized reward outperforms the soft (win rate) variant, in line with our theoretical analysis.}
\label{tab:app:soft-vs-hard}
\end{table}

\section{More Implementation Details}
\label{app:impl}

\subsection{Training Configuration}
\label{app:training-config}

We implement \methodname on top of the VeRL library~\citep{kwon2023efficient}, running on a single node with $8\times$ NVIDIA A800 GPUs.
We use weights $1.0$ for the answer-correctness term, $1.0$ for the format term, and $0.25$ for the hint-injected pairwise term; additional ablations on the pairwise reward configuration are reported in Appendix~\ref{app:soft-vs-hard}.
At each training step we draw $K=8$ hinted rollouts per prompt to estimate $r_{\mathrm{\methodname}}$.
The remaining training hyperparameters are listed in Table~\ref{tab:app:implementation:config}.

\begin{table}[ht]
\centering
\small
\setlength\tabcolsep{2.68pt}
\begin{tabular}{llc}
\toprule
\textbf{Category} & \textbf{Parameter} & \textbf{Value} \\
\midrule
\multirow{6}{*}{General}
  & Advantage estimator           & GRPO  \\
  & Gamma ($\gamma$)              & 1  \\
  & Lambda ($\lambda$)            & 1  \\
  & Batch size                    & 128  \\
  & Max prompt length             & 1024  \\
  & Gradient checkpointing        & Enabled  \\
\midrule
\multirow{10}{*}{Actor}
  & Learning rate                 & $1 \times 10^{-6}$  \\
  & Mini-batch size               & 1024  \\
  & Dynamic batch size            & Enabled  \\
  & KL penalty role               & Loss \\
  & KL loss type                  & Low-variance KL \\
  & KL loss coefficient ($\beta$) & 0.001  \\
  & Entropy coefficient           & 0.001  \\
  & Clip ratio                    & 0.2  \\
  & Gradient clipping             & 1.0  \\
  & Sequence parallel size        & Model-specific  \\
\midrule
\multirow{6}{*}{Rollout}
  & Backend                       & vLLM  \\
  & Tensor model parallel size    & 1  \\
  & Rollouts per sample           & 8  \\
  & Nucleus sampling $p$          & 1.0  \\
  & GPU memory utilization        & 0.6  \\
  & Sampling temperature          & 1.0  \\
\midrule
\multirow{1}{*}{Qwen2.5-7B}
  & Max response length           & 2048  \\
\bottomrule
\end{tabular}
\caption{Training configuration for \methodname. We use VeRL version 0.5.}
\label{tab:app:implementation:config}
\end{table}

\begin{table}[ht]
\centering
\small
\setlength\tabcolsep{0.7pt}
\begin{tabular}{l|cccc|c}
\toprule
\textbf{Model} & \textbf{DeepScaleR} & \textbf{MATH-500} & \textbf{CARP-EN} & \textbf{TheoremQA} & \textbf{Avg.} \\
\midrule
Qwen3$_\mathrm{4B}$   & 27.2 & 49.6 & 46.2 & 29.1 & 38.0 \\
\;+\,SFT            & 39.5 & 60.2 & 48.1 & 31.8 & 44.9 \\
\methodname    & \textbf{62.1} & \textbf{74.6} & \textbf{57.4} & \textbf{51.2} & \textbf{61.3} \\
\bottomrule
\end{tabular}
\caption{Results on Qwen3-4B trained on DeepScalR. \methodname improves both in-domain and out-of-domain accuracy, demonstrating backbone-agnostic gains.}
\label{tab:app:qwen3-4b-math}
\end{table}

\begin{table}[ht]
\centering
\small
\setlength\tabcolsep{4pt}
\begin{tabular}{l|cccc|c}
\toprule
\multirow{2}{*}{\textbf{Model}} & \multirow{2}{*}{\textbf{MedQA}} & \multirow{2}{*}{\textbf{MedMCQA}} & \multicolumn{2}{c|}{\textbf{MMLU-Pro}} & \multirow{2}{*}{\textbf{Avg.}} \\
 & & & Health & Bio & \\
\midrule
Qwen3$_\mathrm{4B}$ & 57.2 & \textbf{62.2} & 38.1 & 50.9 & 52.1 \\
\;+\,SFT  & 51.6 & 49.7 & 46.8 & 72.6 & 55.2 \\
\methodname  & \textbf{58.1} & 53.3 & \textbf{56.7} & \textbf{78.8} & \textbf{61.7} \\
\bottomrule
\end{tabular}
\caption{Results on Qwen3-4B trained on MedQA.}
\label{tab:app:qwen3-4b-med}
\end{table}

\begin{table}[t]
\centering
\small
\renewcommand{\arraystretch}{1.1}
\begin{tabular}{p{3.5cm} p{3.5cm}}
\toprule
\textbf{RL Dataset / Family} & \textbf{Direct Upstream Sources} \\
\midrule
\textbf{SimpleRL-Zoo} & GSM8K (train) + MATH (train) \\
\midrule
\textbf{Big-Math-RL-Verified} & MATH + GSM8K + AMC/AIME + AoPS \\
\midrule
\textbf{DeepScaleR} & AIME (1984–2023) + AMC ($\le 2023$) + Omni-MATH \\
\midrule
\textbf{NuminaMath Series} \newline (OpenR1, Skywork-OR1) & AoPS + MATH + GSM8K + AMC/AIME \\
\bottomrule
\end{tabular}
\caption{Summary of \contamination lineages in widely adopted RL reasoning datasets. These datasets directly source from benchmarks that have been empirically proven to be exposed to modern base models prior to RL training.}
\label{tab:contamination_evidence}
\end{table}

\subsection{Prompts}
\label{app:prompts}

We document the prompt templates used in our pipeline:
the rubric prompt that produces the dimension-wise quality labels for the gold score $r(\cdot)$ (Table~\ref{tab:app:prompt:quality}),
the pairwise-judge template used by both the reward judge $J$ and the PNT analysis (Table~\ref{tab:app:prompt:pairwise}),
and the two rollout templates that materialize the unhinted policy $\pi_\theta$ and the hinted reference $\pi_H$ (Table~\ref{tab:app:prompt:rollout}).

\begin{table*}[ht]
\small
\centering
\begin{tabular}{p{0.96\textwidth}}
\toprule
\textbf{1. Accuracy (Score: 1, 0, -1 or -2)} \newline
\textit{Focus:} Does the final answer match the correct answer? \newline
\textbf{Score 1 (Correct):} The answer is numerically or semantically equivalent to the correct answer.
This includes minor formatting differences, unsimplified fractions, scientific notation, rounding to an acceptable range, or the use of equivalent values (e.g., if the model provides a decimal or fraction and the correct answer is an integer rounded from that decimal, this is acceptable). \newline
\textbf{Score 0 (Incorrect):} The answer is wrong, missing, or fundamentally different. \newline
\textbf{Score -1 (Reference Error):} If the correct answer is incomplete (e.g., a question asks for both maximum and minimum values, but only one value is provided), assign a $-1$ to indicate that the reference data requires review. \\
\midrule
\textbf{2. Knowledge (Score: 1 or 0)} \newline
\textit{Focus:} This dimension checks only for errors in memorized facts (e.g., value of constants, standard textbook formulas). It does not penalize misuse of valid knowledge. \newline
\textbf{Score 1 (Correct):} All external knowledge (such as mathematical formulas, physical constants, legal statutes, medical symptoms, etc.) used is theoretically correct (e.g., stating $E=mc^2$ or $\pi \approx 3.14159$). \newline
\textit{Note 1:} If the model uses a correct formula (e.g., $V = \tfrac{4}{3}\pi r^3$ for the volume of a sphere), even if the formula is not relevant to the problem, this still counts as 1 for Knowledge. We care only about whether the factual knowledge is correct, not whether it is applied in the right context. \newline
\textit{Note 2:} If the model writes the correct formula but calculates it wrong or substitutes the wrong value into the formula, this is still Score 1 for Knowledge. \newline
\textit{Note 3:} If the model hallucinates a specific equation for the problem (e.g., ``We assume $x + y = 10$'') that turns out to be invalid for the scenario, this is Score 1 here. It is an assumption failure rather than a Knowledge error unless it is a universal theorem. \newline
\textbf{Score 0 (Error):} The model hallucinates a false universal truth, such as a non-existent formula or remembering a constant incorrectly (e.g., stating ``The sum of angles in a triangle is 360 degrees'', attributing a historical event to the wrong century). \\
\midrule
\textbf{3. Calculation (Score: 1 or 0)} \newline
\textit{Focus:} Is the arithmetic execution correct? \newline
\textbf{Score 1 (Correct):} All arithmetic calculations are performed correctly based on the numbers present in the reasoning.
\textit{Note:} A wrong final answer does not necessarily indicate a calculation error; it could be due to using the wrong formula or flawed logic. This category focuses only on the correctness of the calculations themselves. Besides, if there is no need for calculations in the problem, then this should be scored as 1. \newline
\textbf{Score 0 (Error):} The model makes calculation errors (e.g., calculating the wrong result number, substituting the wrong value into a correct formula). \\
\midrule
\textbf{4. Logic Properness (Score: 1 or 0)} \newline
\textit{Focus:} This evaluates the logical coherence of the reasoning process. Check whether each step logically follows from the previous ones, and whether all necessary conditions are satisfied to draw the conclusion. \newline
\textbf{Score 1 (Valid):} The reasoning is coherent, and each step logically follows from its premises. \newline
\textbf{Score 0 (Flawed):} When the reasoning involves any of the following: \newline
\textbullet\ \textit{Insufficient conditions:} the evidence is insufficient to draw the conclusion (e.g., concluding a quadrilateral is a square solely because it has 4 right angles, missing the necessary condition of equal side lengths). \newline
\textbullet\ \textit{Self-contradiction:} the model makes logical errors such as contradicting its own earlier reasoning (e.g., first states ``$x>0$'' and later assumes ``$x=-5$''), or draws conclusions that do not logically follow from the premises (e.g., reasoning points to Answer A, but the final conclusion is Answer B). \newline
\textbullet\ \textit{Other logical mistakes.} \\
\midrule
\textbf{5. Assertion Correctness (Score: 1 or 0)} \newline
\textit{Focus:} This evaluates the correctness of assertions made during the reasoning process. Specifically, it checks if the model introduces any assumptions or numbers that appear out of nowhere and lack support. \newline
\textbf{Score 1 (Grounded):} All numbers, variables, and constraints introduced are traceable to the question or previous steps. \newline
\textbf{Score 0 (Hallucinated/Fabricated):} When the reasoning involves any of the following: \newline
\textbullet\ \textit{Fabricated constraints:} the model introduces rules or assumptions that cannot be supported by the question or previous steps (e.g., ``Assume $x$ is an integer'' or ``Assume the triangle is equilateral'' when this was not provided). \newline
\textbullet\ \textit{Magic steps/numbers:} new numbers or results appear without any explanation or calculation (e.g., suddenly stating ``Let's consider a possible solution of 12769'' without showing the calculation or derivation). \newline
\textbullet\ \textit{Reverse engineering:} abruptly stating the correct answer without a sufficient reasoning path. \newline
\textit{Note 1 (vs.\ Knowledge error):} if the model hallucinates a specific equation for the problem (e.g., ``We assume $x + y = 10$'') that turns out to be invalid for the scenario, this is not a Knowledge error but an Assertion error, as the specific equation is not a universal theorem. \newline
\textit{Note 2 (vs.\ Calculation error):} if an incorrect assertion stems from a visible computation/derivation, and the claim is wrong only because the calculation itself is incorrect (e.g., claiming ``$2^5+1$ is divisible by 4'' due to an incorrect evaluation of $(2^5+1)\bmod 4$), this should be treated as a Calculation error, not an Assertion error. However, if the model does not explain where the number/claim comes from at all, and it simply appears out of nowhere, then it is a magic number / fabricated claim, which is an Assertion error. \\
\bottomrule
\end{tabular}
\caption{Quality-evaluation rubric used to produce the dimension-wise gold labels referenced in Appendix~\ref{app:gold-score}. A trace is labeled \emph{good} only if it receives a passing score on every dimension.}
\label{tab:app:prompt:quality}
\end{table*}

\begin{table*}[ht]
\small
\centering
\begin{tabular}{p{0.96\textwidth}}
\toprule
You are an expert evaluator of reasoning chains in the medical domain.
Below are two reasoning traces (Chain-of-Thought) for the same medical/mathmetical problem. \newline
\newline
\textbf{Problem}: \texttt{\{question\}} \newline
\textbf{Trace A}: \texttt{\{trace\_a\}} \newline
\textbf{Trace B}: \texttt{\{trace\_b\}} \newline
\newline
\textbf{Evaluation Criteria.}
Compare the traces based on the following 5 dimensions. A "better" trace is one that exhibits higher rigor and fewer flaws across these categories: \newline

1. Logical Properness \& Coherence: seamless logical flow. Penalize logic jumps or self-contradictions. \newline
2. Assertion Correctness \& Groundedness: no ``magic numbers'' or ``reverse engineering'' to match a target answer. \newline
3. Clinical Reasoning Rigor \& Knowledge Validity: follows established clinical guidelines and does not hallucinate false medical facts. \newline
4. Depth \& Clarity: comprehensive, step-by-step derivation over abrupt or shallow explanations. Penalize unnecessary verbosity. \newline
\newline
\textbf{Task.}
Analyze both traces according to the criteria above. Determine which one demonstrates a more authentic, efficient, and rigorous reasoning process overall. \newline
\newline
Output your final decision in the following format: \newline
\texttt{[[Decision]]: [Trace A is better / Trace B is better / Equal quality]} \\
\bottomrule
\end{tabular}
\caption{Pairwise-judge prompt used by the reward comparator $J$ and the PNT analysis.}
\label{tab:app:prompt:pairwise}
\end{table*}

\begin{table*}[ht]
\small
\centering
\begin{tabular}{p{0.96\textwidth}}
\toprule
\textbf{Unhinted rollout (active policy $\pi_\theta$).} \newline
\texttt{system:} A conversation between User and Assistant. The user asks a question, and the Assistant solves it. The assistant first thinks about the reasoning process in the mind and then provides the user with the answer. The reasoning process and answer are enclosed within \texttt{<think> </think>} and \texttt{<answer> </answer>} tags, respectively, i.e., \texttt{<think>} reasoning process here \texttt{</think>} \texttt{<answer>} answer here \texttt{</answer>}. \newline
\texttt{user:} \texttt{\{question\}} \newline
\texttt{assistant:} \texttt{<think>} \\
\midrule
\textbf{Hinted rollout (contaminated reference $\pi_H$).} \newline
\texttt{system:} A conversation between User and Assistant. The user asks a question, and the Assistant solves it. The assistant provides the final answer first, then follows up with a comprehensive reasoning process. The answer and reasoning process are enclosed within \texttt{<answer> </answer>} and \texttt{<think> </think>} tags, respectively, i.e., \texttt{<answer>}answer here\texttt{</answer><think>}reasoning process here\texttt{</think>}. \newline
\texttt{user:} \texttt{\{question\}} \newline
\texttt{assistant:} \texttt{<answer>} \texttt{\{the correct answer\}} \texttt{</answer><think>} \\
\bottomrule
\end{tabular}
\caption{The two rollout templates. The unhinted template (top) elicits a reasoning-first generation that defines the active policy $\pi_\theta$. The hinted template (bottom) front-loads the ground-truth answer so the subsequent reasoning becomes a post-hoc rationalization, materializing the contaminated reference $\pi_C$ used by the pairwise reward $r_{\mathrm{\methodname}}$.}
\label{tab:app:prompt:rollout}
\end{table*}

\subsection{Data Use}
Our use of the datasets is consistent with their intended purposes and complies with their respective licenses.

\section{Additional Experimental Results}
\label{app:more-exp}    

\subsection{Generalization Across Backbones: Qwen3-4B}
\label{app:qwen3-4b}

To verify that the gains of \methodname are not tied to a specific backbone, we replicate the main experiment on Qwen3-4B. Table~\ref{tab:app:qwen3-4b-math} and Table~\ref{tab:app:qwen3-4b-med} reports in-domain and out-of-domain (TheoremQA) accuracy. \methodname consistently outperforms both the original instruction-tuned model and the SFT-then-standard-RL baseline, confirming that the contamination-resistant signal transfers across model scales and architectures.

\subsection{Hard vs.\ Soft Pairwise Reward}
\label{app:soft-vs-hard}

The main results use the binarized (hard) variant of $r_{\mathrm{\methodname}}$, in which the Monte Carlo win-rate is thresholded at $0.5$. We compare this against a continuous (soft) variant that uses the raw win-rate as the reward, and a hard variant with different weight.

As shown in Table~\ref{tab:app:soft-vs-hard}, the hard variant outperforms the soft variant at matched reward scale. This is consistent with the theoretical analysis: Lemma~\ref{lem:tv} attains its supremum precisely at the indicator witness $T^\star = \mathbf{1}_{A^\star}$, so a $\{0,1\}$-valued reward is the one that tightens the lower bound in Eq.~\eqref{eq:lb} on the underlying divergence.



\section{Evidence of \Contamination in Common RL Reasoning Datasets}
\label{app:data_overlap}

A critical premise of our work is that many widely adopted reasoning datasets utilized in recent RL post-training pipelines inherently suffer from \contamination. Specifically, the upstream benchmarks used to construct these RL datasets have already been ingested by modern LLMs during their pre-training or supervised fine-tuning (SFT) phases. In this section, we systematically trace four reliable lineages of data overlap, demonstrating how state-of-the-art RLVR (Reinforcement Learning with Verifiable Rewards) datasets inherit this contamination. A high-level summary is presented in Table~\ref{tab:contamination_evidence}.

The datasets primarily trace back to four reliable lineages of overlap. \textbf{Chain 1: SimpleRL-Zoo}~\cite{zeng2025simplerlzooinvestigatingtamingzero}. This dataset directly utilizes the official training splits of GSM8K~\cite{cobbe2021gsm8k} and MATH~\cite{hendrycksmath2021}. The contamination of these upstream benchmarks is extensively documented, with models like Qwen2.5-Math-7B exhibiting up to a 54.6\% exact-match regurgitation rate on MATH prompts~\citep{wu2025reasoningmemorizationunreliableresults}. \textbf{Chain 2: Big-Math-RL-Verified}~\cite{albalak2025bigmathlargescalehighqualitymath}. Comprising over 250k instances, it explicitly includes subsets of MATH, GSM8K, and AMC/AIME. \textbf{Chain 3: DeepScaleR}~\cite{deepscaler2025}. This dataset relies heavily on historical AIME and AMC problems. \textbf{Chain 4: The NuminaMath Ecosystem}~\cite{numina_math_datasets}. Datasets such as OpenR1-Math-220k and Skywork-OR1~\cite{skywork-or1-2025} rely on NuminaMath as their foundational data source, a dataset that is itself pervasively utilized as a pre-RL training corpus~\cite{mahdavi2025leveragingonlineolympiadlevelmath}.
\section{Case Study}
\label{app:case-study}

To complement the aggregate metrics with a qualitative view, we present a representative prompt from the MedQA validation set. On this instance, the SFT-then-standard-RL baseline produces a reasoning trace riddled with severe hallucinations to rationalize a memorized correct answer, while \methodname produces a reliable, faithful derivation.

\begin{table*}[htbp]
\centering
\small
\renewcommand{\arraystretch}{1}
\begin{tabular}{p{0.96\textwidth}}
\toprule
\textbf{Prompt.} A 4-year-old girl from a recently immigrated family presents to the emergency department with episodes of severe coughing lasting up to several minutes followed by vomiting. She had a low grade fever and runny nose over the last 2 weeks but these coughing episodes just began one day prior to presentation. A complete blood count shows a lymphocytic infiltrate and Gram stain reveals a gram-negative coccobacillus. The emergency department physician explains that this organism causes disease by toxin-mediated inactivation of an inhibitory signaling molecule. Which of the following could be used to culture the most likely cause of this disorder? \\
Options: 
(A) Charcoal yeast with iron and cysteine 
(B) Eaton agar 
(C) Loffler medium 
(D) Regan-Lowe medium 
(E) Thayer-Martin agar \\
\midrule

\textbf{SFT-then-standard-RL output} (abbreviated) \newline
\texttt{<think>} The most likely cause of this patient's severe coughing and vomiting is \textit{Pseudomonas aeruginosa}, a gram-negative coccobacillus. The pathophysiology of \textit{Pseudomonas aeruginosa} involves the production of a potent exotoxin, pseudomonas exotoxin A (PE), which is a type II protein-toxin. PE is a preformed, heat-stable protein that is secreted by the bacterium. Once released into the host, PE is cleaved by host trypsin to become active. The active form of PE is a highly toxic, highly immunogenic protein that acts as a cytotoxin. It is best known for its ability to inactivate the host's inhibitory signaling molecule, the inhibitory G-protein alpha subunit (Galpha-i). By deactivating Galpha-i, the toxin prevents the normal negative feedback of the cyclic AMP (cAMP) signaling pathway, leading to an abnormally high concentration of cAMP in the host cell. This elevated cAMP promotes the expression of virulence genes, such as those encoding for exotoxins and biofilm formation, and enhances the bacterium's ability to colonize and multiply in the host's respiratory tract and other mucosal surfaces.

\textit{Pseudomonas aeruginosa} is a highly opportunistic pathogen that commonly causes severe infections in immunocompromised or chronically ill patients, as well as in patients with underlying lung diseases such as cystic fibrosis. It is a non-fermenting, oxidase-positive, and catalase-positive organism. To culture this organism, a selective medium containing specific growth factors is required. Regan-Lowe medium is a specialized culture medium used for the selective isolation of \textit{Pseudomonas} species. It is a nutrient-rich medium that contains iron, cysteine, and vitamins, which are essential for the growth of \textit{Pseudomonas}. The addition of iron and cysteine provides the necessary sulfur and nitrogen sources for the synthesis of the bacterium's characteristic, highly mucoid, blue-green pigmentation. \texttt{</think>} \texttt{<answer>} D \texttt{</answer>} \\
\midrule

\textbf{\methodname output} (abbreviated) \newline
\texttt{<think>} The clinical presentation of severe, prolonged coughing followed by vomiting, combined with a lymphocytic leukocytosis and a Gram stain revealing a gram-negative coccobacillus, is highly characteristic of \textit{Bordetella pertussis}, the causative agent of whooping cough (pertussis). 

\textit{Bordetella pertussis} is a highly specialized, highly virulent Gram-negative bacterium that induces its pathogenesis through the production of two major exotoxins: pertussis toxin (also known as pertactin) and filamentous hemagglutinin. Pertussis toxin is a highly potent ADP-ribosylating toxin that targets the G$\alpha$i subunit of Gi-coupled G-protein receptors. By inactivating this inhibitory G-protein, the toxin prevents the normal downregulation of adenylate cyclase. This leads to a massive increase in intracellular cAMP, which subsequently activates protein kinase A (PKA). PKA drives the excessive, paroxysmal coughing characteristic of the disease by stimulating the cough center in the medulla oblongata and inhibiting the normal relaxation of the laryngeal muscles. 

The most reliable method for culturing \textit{B. pertussis} is the Regan-Lowe medium. This specialized agar is heavily enriched with vitamins, iron, and cysteine, which are critical for the bacterium's growth. The medium also contains bile salts and crystal violet, which inhibit the growth of other competing organisms. Because \textit{B. pertussis} is highly fastidious and grows very slowly (with a doubling time of approximately 16 hours), it requires specific growth conditions. Therefore, Regan-Lowe medium is the preferred culture medium for isolating this organism. \texttt{</think>} \texttt{<answer>} D \texttt{</answer>} \\
\bottomrule
\end{tabular}
\caption{A model response example from different methods in MedQA validation set.}
\label{tab:app:case-study}
\end{table*}

\paragraph{Analysis.} 
As illustrated in Table~\ref{tab:app:case-study}, the baseline RL model falls into a severe hallucination trap driven by shortcut exploitation. Despite successfully arriving at the correct final option (D), its underlying reasoning is entirely unreliable. 
The model explicitly misidentifies the pathogen as \textit{Pseudomonas aeruginosa}, yet forcibly aligns it with the clinical vignette by attributing the distinct pathophysiological mechanism of \textit{Bordetella pertussis} (the inactivation of the G$\alpha$i protein and subsequent cAMP elevation) to \textit{Pseudomonas} Exotoxin A. Moreover, it explicitly fabricates the claim that Regan-Lowe medium---the definitive culture for \textit{B. pertussis}---is used to selectively isolate \textit{Pseudomonas}. 

In contrast, \methodname exhibits faithful, forward-chaining logical deduction: accurately diagnosing \textit{B. pertussis}, explaining its true toxin mechanism, and logically deriving Regan-Lowe medium as the correct clinical decision.

\section{Limitations}

\paragraph{Scope of Hint Injection.} 
Currently, our framework exclusively employs the correct answers for hint injection. 
However, injecting \textit{incorrect} answers could further elicit diverse reasoning failures, 
allowing the pairwise reward model to capture finer-grained preference signals. 
For instance, it could help the judge distinguish between minor calculation slips and severe logical hallucinations among two incorrect traces, 
thereby providing a denser and more informative signal to boost RL optimization efficiency. 
We leave the exploration of incorrect-hint injection as a promising avenue for future work.

\paragraph{Utilization of Rollout Traces.} 
In our current implementation, the standard and hint-injected rollouts are generated independently from scratch. 
A more efficient and synergistic approach would be to enforce prefix-sharing or apply token-level perturbations, ensuring both traces share the identical initial reasoning path before divergence. 
This alignment would make the paired traces strictly comparable, substantially reducing the capability requirements for the judge model. 
Furthermore, it could naturally facilitate self-correction mechanisms~\cite{kumar2024traininglanguagemodelsselfcorrect}. 
Optimizing the utilization of rollout trajectories to unlock such self-evolution remains an open challenge.

\paragraph{Computational Overhead.} 
Although \methodname utilizes a relatively small 4B LLM as the pairwise reward model, running an additional model concurrently during RL inevitably introduces computational overhead. A faster alternative could involve replacing the LLM judge with a lightweight discriminator. However, training such a discriminator typically requires additional labeled preference data, which compromises the data-efficiency of our current framework. We therefore leave the development of a lightweight, label-efficient judge to future research.


\end{document}